\documentclass[10pt,twocolumn,letterpaper]{article}

\usepackage{cvpr}
\usepackage{times}
\usepackage{epsfig}
\usepackage{graphicx}
\usepackage{amsmath}
\usepackage{amssymb}
\usepackage{euscript}
\usepackage{multirow}

\usepackage[breaklinks=true,bookmarks=false]{hyperref}

\cvprfinalcopy 

\def\cvprPaperID{****} 

\begin{document}

\title{GeoGraph: Learning graph-based multi-view object detection with geometric cues end-to-end}

\author{Ahmed Samy Nassar\textsuperscript{1,2}, Stefano D'Aronco\textsuperscript{2}, {S{\'e}bastien Lef{\`e}vre}\textsuperscript{1},  
Jan D. Wegner\textsuperscript{2}\\
\textsuperscript{1}IRISA, Universit{\'e} Bretagne Sud\\
\textsuperscript{2}EcoVision Lab, Photogrammetry and Remote Sensing group, ETH Zurich\\
}

\maketitle
\begin{abstract}
In this paper we propose an end-to-end learnable approach that detects static urban objects from multiple views, re-identifies instances, and finally assigns a geographic position per object. Our method relies on a Graph Neural Network (GNN) to, detect all objects and output their geographic positions given images and approximate camera poses as input. Our GNN simultaneously models relative pose and image evidence, and is further able to deal with an arbitrary number of input views. Our method is robust to occlusion, with similar appearance of neighboring objects, and severe changes in viewpoints by jointly reasoning about visual image appearance and relative pose. Experimental evaluation on two challenging, large-scale datasets and comparison with state-of-the-art methods show significant and systematic improvements both in accuracy and efficiency, with 2-6\% gain in detection and re-ID average precision as well as 8x reduction of training time. 

\end{abstract}

\section{Introduction}
We present an end-to-end trainable multi-view object detection and re-identification approach centered on Graph Neural Networks (GNN). Unlike images, much data is not structured on a grid but naturally follows a graph structure. GNNs apply directly to graphs and thus their applications vary over many disciplines like predicting molecular properties for chemical compounds \cite{gilmer2017neural,li2018learning} and proteins \cite{fout2017protein}, social influence prediction \cite{qiu2018deepinf}, object tracking \cite{braso2019learning,gao2019graph}, or detection of fake news~\cite{monti2019fake}. Here, we propose to solve the problem of multi-view detection and re-identification of static objects in urban scenes using Graph Neural Networks. Given a set of ground-level images with coarse relative pose information, we detect, re-identify and finally assign geographic coordinates to thousands of urban objects with an end-to-end learnable approach.


Maintaining complete and accurate maps of urban objects is essential for a wide range of applications like autonomous driving, or maintenance of infrastructure by local municipalities. Despite much research in this field \cite{wegner2016cataloging,krylov2018automatic,zhang2019automated,nassar2019simultaneous}, updating maps is often still carried out via field surveys, which is a time-consuming and costly process. Here, we propose to accomplish this task by leveraging publicly available imagery that comes with coarse camera pose information. 


What makes this task challenging is the relatively poor image quality of street-level panoramas (e.g., image stitching artefacts, motion blur) or dash cam image sequences (e.g., motion blur, narrow field of view) compared to data acquired through dedicated mobile mapping campaigns. Basically, wide baselines between consecutive acquisitions, and inaccurate camera poses information hinder establishing dense pixel-level correspondences between images. We thus propose to integrate image evidence and pose information into a single, end-to-end trainable neural network that uses images and coarse poses as input and outputs the geo-location of each distinct object in the scene. In contrast to a rigorous structure-from-motion approach, our method learns the joint distribution over different warping functions of the same object instance across multiple views together with the relative pose information. Unlike recent research in this domain, e.g., ~\cite{wegner2016cataloging,lefevre2017toward,branson2018google,krylov2018automatic}, our method employs an end-to-end approach that helps to jointly learn features to carry out the detection, re-identification, and geo-localization tasks. And differently to other end-to-end works such as~\cite{nassar2019simultaneous}, our approach for re-identification is based on a GNN that enables the use of multiple views (2+) and is much more computationally efficient in comparison to a siamese approach.

\begin{figure*}
\begin{center}
\includegraphics[scale=0.4]{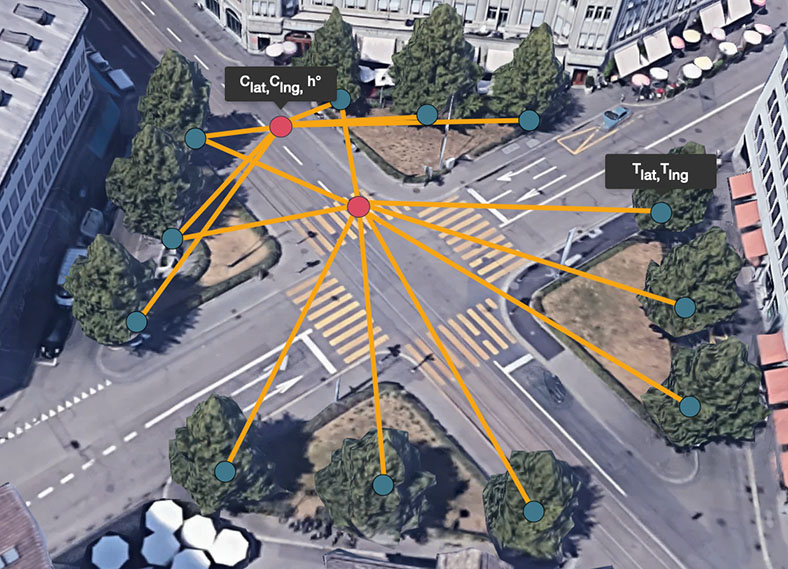}
\end{center}
  \caption{Illustration of our multi-view scenario. Red circle: Camera acquisition location. Green circle: target object to be detected. Orange line: distance between camera and object.}
\label{fig:setup}
\end{figure*}

As illustrated in Fig.~\ref{fig:setup}, our method works as follows: using a set of street-level images that come with coarse camera pose information as input, an object detector predicts several object bounding boxes for each of the views.
We then construct a fully connected graph connecting all the object instances across the images. Next, the graph is fed through a GNN whose goal is to separate it into multiple disconnected sub-graphs, each representing a distinct object in the scene. The GNN has access to both image evidence and coarse pose information, so that it can learn to merge geometric view information with the corresponding object features from different viewpoints and to ensure high quality object re-identification. Finally, once that the distinct objects are re-identified, the proposed end-to-end architecture estimates their geo-location.

\paragraph{Our contributions are:} \emph{i}) an efficient end-to-end, multi-view detector for static objects \emph{ii}) that implements a novel method for incorporating graphs inside any anchor-based object detector. We further \emph{iii}) formulate a GNN approach that jointly uses coarse relative camera pose information and image evidence to detect distinct objects in the scene. We validate our method experimentally on two different datasets of street-level panoramas and dash cam image sequences. Our GNN formulation for multi-view object detection and instance re-identification outperforms existing methods while being much more computationally efficient. 



\section{Related Work}
Our proposed method is related to many different research topics in computer vision like pose estimation, urban object detection, and instance re-identification. A full review is beyond the scope of this paper and we rather provide here some representative works for each different topic and highlight the differences with our proposal.

\textbf{Urban object detection} from ground-level images is an application closely related to our paper. In~\cite{wegner2016cataloging,branson2018google}, the authors propose a method to detect and geo-locate street-trees from Google street-view panorama and aerial images with a hierarchical workflow. Trees are first detected in all images, detections are then projected to geographic coordinates. The detection scores are back-projected into images and re-evaluated, finally a conditional random field integrates all image evidence with other learned priors. In~\cite{zhang2018using} a method is proposed to detect and geo-locate poles in Google street-view panoramas using object detectors along with a modified brute-force line-of-bearing approach to estimate pole locations. Authors in~\cite{krylov2018object} perform a semantic segmentation of images and estimate monocular depth before feeding both sources of evidence into a Markov random field to geo-locate traffic signs. \cite{zhang2019automated} detects road objects from ground level imagery and places them into the right location using semantic segmentation and a topological binary tree. All previously mentioned methods have in common that they propose hierarchical, multi-step workflows where pose and image evidence are treated separately unlike ours, which models them jointly. The most similar work to ours is~\cite{nassar2019simultaneous}, which proposes to jointly leverage relative camera pose information and image evidence as an end-to-end trainable siamese CNN. Here, we propose to formulate the problem as a graph neural network, which, intuitively, better represents the underlying data structure of multi-view object detection. Our GNN provides greater flexibility to add an arbitrary number of views (as opposed to a siamese CNN), is computationally drastically less costly, and achieves significantly better quantitative results.
A large body of literature addresses urban object detection from an autonomous driving perspective with various existing public benchmark datasets like famous KITTI~\cite{geiger2013vision}, CityScapes~\cite{cordts2016cityscapes}, or Mapillary~\cite{neuhold2017mapillary,ma2020state}. As opposed to our case, in this scenario dense image sequences are acquired with minor viewpoint changes in driving direction with forward facing cameras and often relative pose information of good accuracy, enabling object detection and re-identification across views~\cite{chen2017multi,8594049,zhao2018object}. In our setup, relative pose information is coarse and the viewpoints among the images might be remarkably different, i.e., large baseline between the cameras, making the correspondences matching a much harder task.

\textbf{Learning to predict camera poses} using deep learning was made popular by PoseNet~\cite{kendall2015posenet} and still motivates various studies~\cite{hideo2017,en2018,xiang2018}. Estimating a human hand's appearance from any viewpoint can be achieved by coupling pose with image content \cite{poier2018learning}.
Full human pose estimation is another task that benefits from combined reasoning across pose and scene content, for instance in~\cite{luvizon20182d}, authors employ a multi-task CNN to estimate pose and recognize action. We rely here on public imagery without fine-grained camera pose information, which requires a different approach.

\textbf{Multiple object tracking (MOT) and person re-identification} is related to our setting, but significantly differs in that objects are moving while cameras are usually fixed. Again, many deep learning-based solutions have been reported, usually employing a siamese CNN~\cite{li2014deepreid}, as it is an effective technique to measure similarity between image patches. In~\cite{leal2016learning}, for instance, authors propose to learn features using a siamese CNN for multi-modal inputs (images and optical flow maps). In~\cite{wang2016joint}, a siamese CNN and temporally constrained metrics are jointly learned to create a tracklet affinity model.~\cite{sadeghian2017tracking} uses a combination of CNNs and recurrent neural networks (RNN) in order to match pairs of detections.
Authors in \cite{xiao2019ian}, instead, solve the  re-identification problem with a so-called center loss that tries to minimize the distance between candidate boxes in the feature space.
A work that is closely related to ours is~\cite{braso2019learning}. Authors formulate their MOT for person re-identification into a graph setup. The graph is composed of nodes that hold CNN features of the image crops of the persons over time, with edges created between all these instances. A message passing network is then used to propagate the node features throughout the graph. Similar to our work, the edges between these nodes are then classified to re-identify the person. This paper however focuses on a single camera setup, whereas in our case we need to re-identify objects from different views. 


\textbf{Graph neural networks (GNN)} naturally adapt to non-grid structured data like molecules, social networks, point clouds, or road networks. 
Graph convolutional networks (GCNs), as introduced by~\cite{bruna2013spectral}, originally proposed a convolutional approach on spectral graphs, which was further extended in~\cite{defferrard2016convolutional,kipf2016semi,8521593}. Another line of research investigated GCNs in the spatial domain~\cite{atwood2016diffusion,niepert2016learning,gilmer2017neural,monti2017geometric,fey2018splinecnn}, where spatial neighborhoods of nodes inside graphs are convolved. 
In order to make processing on large graphs more efficient, recent methods pool nodes in order to perform subgraph-level classification~\cite{hamilton2017inductive,velivckovic2017graph}. These methods are described as fixed-pooling methods because they are based on the graph topology. Another idea is pooling nodes into a coarser representation based on weighted aggregations that are learned from the graph directly~\cite{ying2018hierarchical,gao2019graph,diehl2019edge}.
In our approach, a GNN network is used to capitalize on the data structure by classifying edges between nodes to find the correspondence. We use the spatial convolution operator GraphConv \cite{morris2019weisfeiler} since spatial-based convolutions have proven to be more efficient, and to generalize better on other data~\cite{hamilton2017inductive,wu2019comprehensive}.



\section{Method}
\begin{figure*}
\begin{center}
\includegraphics[width=\textwidth]{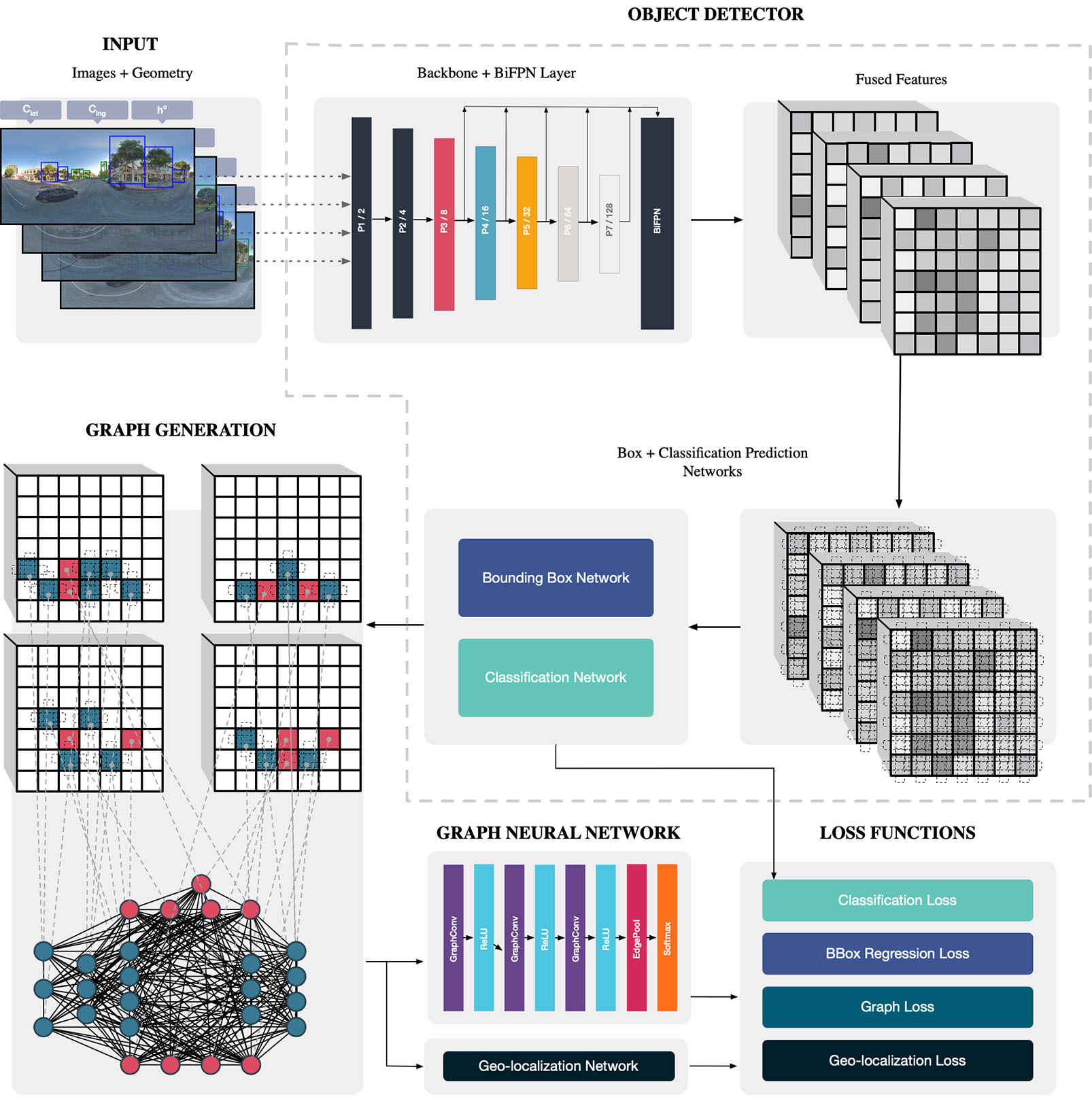}
\end{center}
  \caption{\textbf{Architecture of GeoGraph}. A batch of images from multiple views and camera metadata information $M$ pertaining to them are passed through the backbone network (EfficientNet) and the multi-scale feature aggregator (BiFPN) of the object detector that provides different levels of features. Anchors are then generated across the feature layers, and passed through two sub-networks to provide classification and box predictions. Based on the IoU of the ground truth with the anchors we select the positive and negative anchors. The features of these anchors are used to generate a dense fully connected graph. The graph is then fed to a GNN to predict if the nodes are corresponding by classifying their respective edge. In Parallel, the regressed bounding boxes of the positive anchors are passed to the Geo-Localization Network to regress the geo-coordinate.}
  \clearpage
\label{fig:workflow}
\end{figure*}

We now describe in detail the proposed method. It is convenient to think of our architecture as composed of three stages: object detector, a GNN object re-identification and a geo-localization predictor, see Fig.~\ref{fig:workflow}. Note that, though we can see the model as made of three stages, the training is still conducted in an end-to-end manner, as the all the different parts support back-propagation.


\subsection{Object Detection} 

In general, any state-of-the-art object detector could be used here. We choose EfficientDet~\cite{tan2019efficientdet} as it can provide state-of-the-art accuracy performance with a moderate amount of parameters and computations. 
The benefit of using an anchor based detector, is that our method can be adapted to other state-of-the-art methods~\cite{he2017mask,liu2019cbnet} that also use an anchor based solution. The whole pipeline could, however, easily be adapted to work with anchor-free object detectors if desired.

The input to the object detector is a set of multiple images, which represent a scene, alongside with their metadata $ M = \{ C_{lat}^*,C_{lng}^*, h^\circ \} $ where $C_{lat}^*,C_{lng}^*, h^\circ$ represent the cameras's latitude, longitude and heading angle respectively, which corresponds to the location in the 3D world of the cameras.
Similar to most object detectors, we feed the images through the \emph{backbone}, which is in our case an ImageNet-pretrained CNN, to extract features. The features maps are then fed to both a classifier network and a regression network in order to predict the class of the objects and regress the coordinates of its bounding box, respectively.
During the training phase we obviously have access to the bounding box ground truth with objects IDs of the annotated instances. 

We use the Focal Loss introduced in~\cite{lin2017focal} to train the object detection classifier. This loss was selected as it helps to handle the class imbalance between positive and negative samples. In order to calculate the detection losses, we measure Intersection Over Union (IoU) and select the anchors that have the best overlap with our ground truth bounding boxes. This task is achieved by a threshold that splits the anchors into positive and negative proposals; both proposal will then be used in the next stage of our architecture (see Sec.~\ref{sec:reid}). As for the object bounding box regression, we adopt a smooth-$L_1$ loss~\cite{girshick2015fast}.


\subsection{Object Re-identification} \label{sec:reid}
After detecting the objects across the different views, we need a method that is able to re-identify and recover all the distinct objects that appear of the scene. In this case both the number of input detections, and the number of distinct objects that occur in the scene, are not fixed and vary across different scene instances. This irregularity among the different instances poses some modelling challenges, which, fortunately, can be overcome by using graph. Indeed, since graphs are usually used to represent irregular data, most of the graphs algorithms are suited to deal with a non-fixed number of nodes. As a result, we can map the object detections as nodes of a graph, and then use graph methods, which can handle graphs of different sizes, to perform object re-identification. In the following, we first define how the graph is created from the object detections, and then we describe how it is used to carry out the re-identification task. 

\subsubsection*{Graph Generation} \label{graphgen}

We generate a graph $ \mathcal{G} = (V, E) $ where $V$  represents the set of $N$ nodes, and $E$ the set of edges connecting the nodes. From each view in the scene, we extract as many nodes as there are features vectors located inside the anchors proposed by the object detector. We then think of each node $v$ as containing the associated feature vector extracted from the feature map to which we concatenate the coarse pose information of the image and the predicted bounding box values. 
The features selected from the feature map include all the ones contained in the positive and negative anchors predicted by the object detector. We then connect all the nodes in the graph to each other, building essentially an undirected fully connected graph.

During the training phase we build a second undirected graph, $\mathcal{G}_{\text{gt}}$, which contains the same nodes as $\mathcal{G}$ but with edges encoding the identification information.
Using the groundtruth annotations, we set $e_{ij}=1$ if nodes $i$ and $j$ belong to the same object, regardless of which images the nodes $i$ and $j$ are associated to. Otherwise, we set $e_{ij}=0$, meaning that the nodes are disconnected. We basically connect each pair of nodes only if they come from the same object.
Intuitively, it is convenient to think of $\mathcal{G}_{\text{gt}}$ as made of a set of disconnected sub-graph components, each representing a single individual object in the scene. 

\subsubsection*{Graph Neural Network.}\label{method:gnn}

At this stage we are provided with an input graph $\mathcal{G}$, which is a fully connected graph among all the feature cells of all the objects detected in the different views. 
The goal here is to train a GNN that receives as input $\mathcal{G}$, and disentangles the nodes of the graph that belong to different objects, i.e., the GNN should recover $\mathcal{G}_{\text{gt}}$.

We compose our GNN out of 3 GraphConv~\cite{morris2019weisfeiler} layers with a ReLU activation after each convolution. The GraphConv uses message passing to aggregate information from the neighborhood of $i$, denoted as $\mathcal{N}_{i}$, to update the feature representation $H$  by:

\begin{multline}
H^{(k+1)}(v)= \\ f_{1}^{W_{1}^{(k)}}\left(H^{(k)}(v), f_{2}^{W_{2}^{(k)}}\left(\left\{H^{(k)}(w) | w \in \mathcal{N}(v)\right\}\right)\right),
\end{multline}

where $ k $ represents the layer of the GNN, $ W^{k} $ are the trainable/learned weights, $f_{2}^{W_{2}^{(k)}} $ is the aggregation function of $ \mathcal{N}(v) $, and $f_{1}^{W_{1}^{(k)}}$ merges the neighborhood features. 
At this point in our network, we insert a dropout \cite{srivastava2014dropout} layer for regularization. 
The output feature representations of the nodes are then passed to a modified or stripped down EdgePooling layer~\cite{diehl2019edge} for the edge classification. This operation consists in concatenating the features of all the neighboring nodes in the graph, and passing them through a linear transformation followed by a sigmoid non-linearity:
\begin{equation}
s_{i j} = \sigma \left( W \cdot\left(H_{i} \| H_{j}\right)+b \right), 
\end{equation}
where $\sigma()$ denotes the sigmoid function and $\|$ operator denotes a concatenation operation. $s_{i j}$ represents the probability for the nodes $i$ and $j$ to belong to the same object in the scene. In this case the groundtruth value for each edge comes from $\mathcal{G}_{\text{gt}}$ and, as for the object detection, we train this classifier using focal loss to accommodate for dataset imbalance.

It is worth noting that by relying on a graph formulation we are able to effortlessly deal with a varying number of distinct objects in the scenes. In fact, as we simply aim at disentangling the graph to separate the objects, the method is actually oblivious of the number of distinct objects in the scene, and it is able separate the graph in any number of disconnected components. 


\subsection{Geo-localization}\label{method:geolocalization} 
We estimate the geo-coordinates of the identified objects similarly to~\cite{wegner2016cataloging,nassar2019simultaneous}. The regressed bounding boxes values are projected to real world geographic coordinates by taking advantage of the camera information that is coupled with the image. In order to perform this operation we further assume that the terrain is locally flat. By using the projection equations Eq.~\eqref{eq:streetview_enu}-\eqref{eq:streetview_geo2pix} we are able to map the object bounding box pixel locations $x$ and $y$ in East, North, Up (ENU) coordinates $e_x,e_y,e_z$ and secondly recover the position of the object in the real world $O_{lat}, O_{lng}$.

\begin{multline}
\label{eq:streetview_enu}
(e_x,e_y,e_z) = \bigl( R \cos[\mathrm{C}_{lat}]\sin[\mathrm{O}_{lng}-\mathrm{C}_{lat}], \\ R\sin[\mathrm{O}_{lat}-\mathrm{C}_{lat}],
-\mathrm{h^\circ} \ \bigr)
\end{multline}


\begin{equation}
\label{eq:streetview_geo2pix}
\begin{split}
 x =& \left(\pi + \arctan(e_x, e_y) - \mathrm{h^\circ}\right)W/{2\pi} \\
 y =& \left(\pi/2 - \arctan(C_{h}, z) \right) H/{\pi}
\end{split}
\end{equation}

where $R$ denotes the Earth's radius, $W$ and $H$ are the image's width and height, $C_{h}$ denotes camera's height and $z=\sqrt{e_x^2+e_y^2}$ is an estimate of the object's distance from the camera.
In order to improve the geo-localization accuracy we refine the predictions by feeding the geo-coordinates computed using Eq.~\eqref{eq:streetview_enu}-\eqref{eq:streetview_geo2pix} through a neural network. The geo-localization network is trained using the groundtruth through regression to obtain the object's geo-coordinate using a Mean Square Error (MSE) loss.

%

\subsection{Inference operations}\label{sec:inference}

At inference time, the operations carried out in the whole pipeline are slightly different. First, the object detector generates the proposals for the anchors with classification confidences and bounding boxes.
A threshold is then applied to the classification score to select only the detection proposals with a high classification confidence. Afterwards, Non Maximum Suppression (NMS) is used to filter further proposals and reduce redundancy.

With the remaining proposals, we create a fully connected graph as described before, with each node representing a feature vector contained inside the proposals, at which we concatenate the image's camera metadata information. The generated graph is then fed to the GNN network for edge classification.
Finally, we classify the edges scores by applying a threshold on them with a decision boundary of $0.5$. 
At this point, the obtained graph is supposed to be made of several disconnected components, one for each of the distinct objects in the scene. We remove sequentially each set of connected components until no nodes are left.

We finally compute the geo-location of the identified objects by utilizing the camera metadata information and the location of the bounding box in the image for all the views where the identified object appears. All the objects geo-coordinates computed from each different view are then separately refined with the geo-localization network and, finally, averaged to obtain the final prediction.

\section{Experiments}

\subsection{Datasets}\label{sec:datasets}

\begin{figure}[!htbp]
\centering
\includegraphics[scale=0.2]{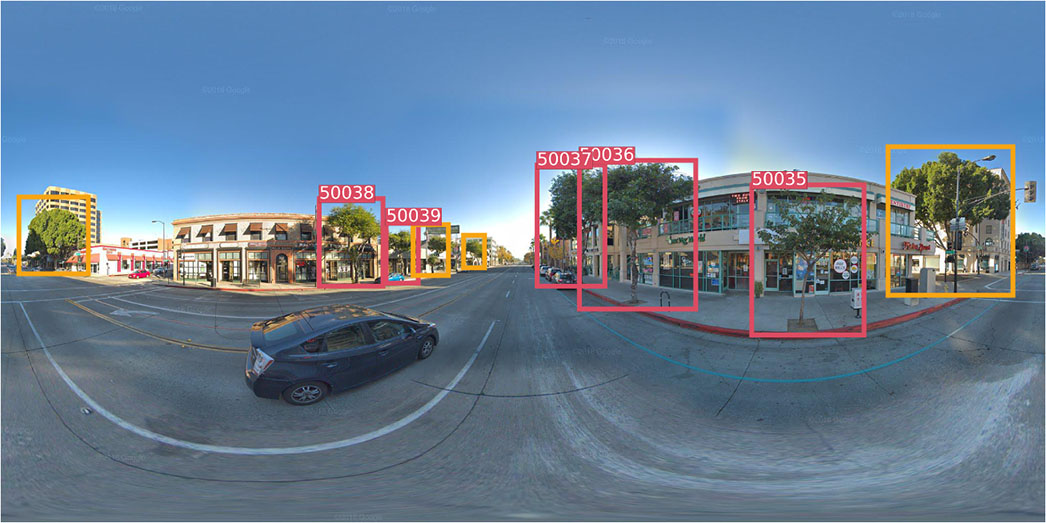}\quad
\includegraphics[scale=0.2]{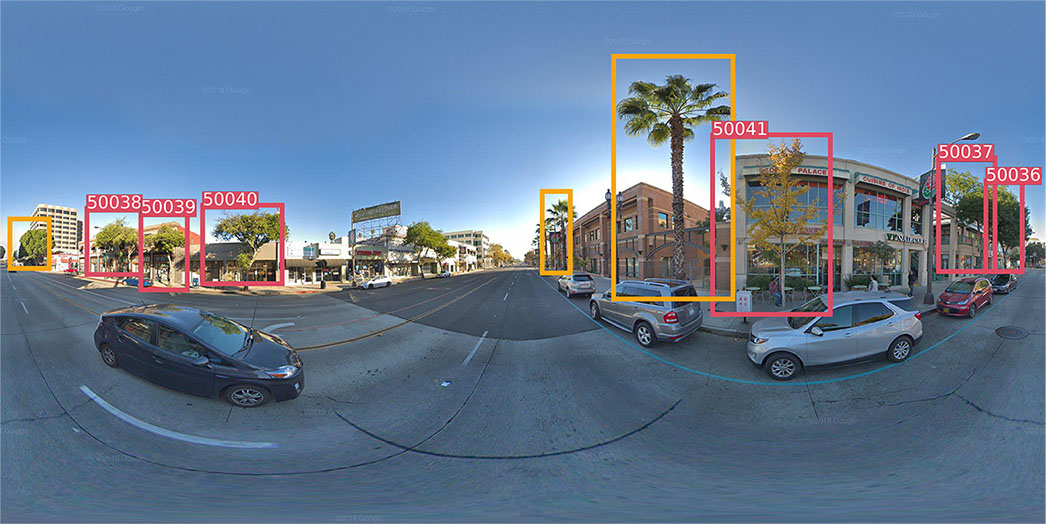}\quad
\medskip
\includegraphics[scale=0.2]{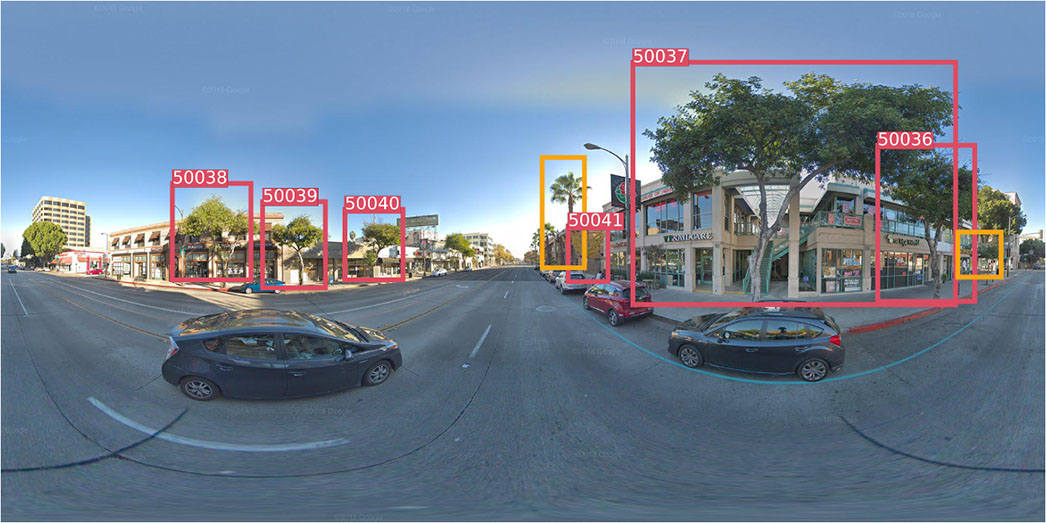}\quad
\includegraphics[scale=0.2]{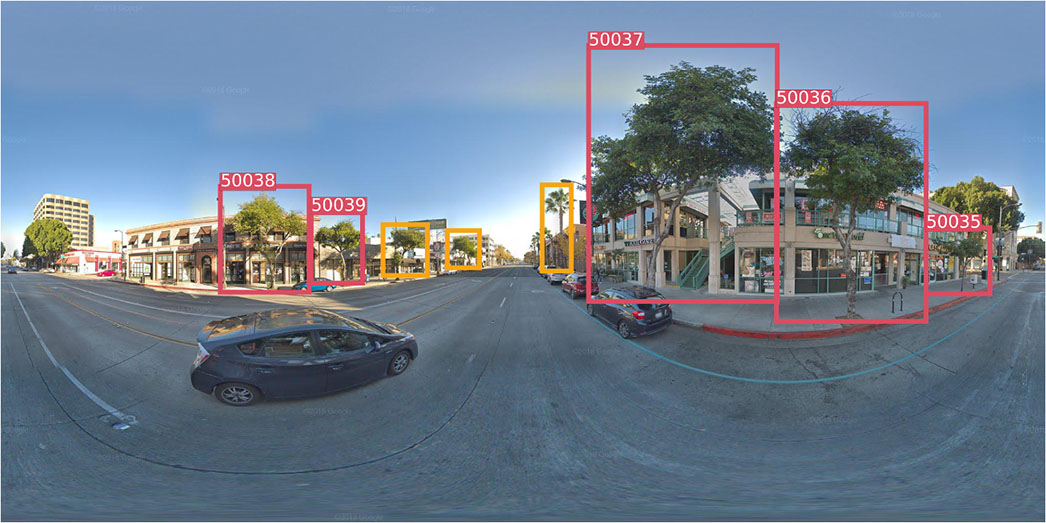}\quad
\caption{Subset of the Pasadena dataset highlighting challenges to be faced by multi-view object detection, re-identification and geo-localization. The trees are in close proximity and share very similar appearances due to their same species and size (trees on the curb are usually planted together at the same time). Some objects are annotated with their instance label (red), some without (orange).}
\label{fig:pasadena_sample}
\end{figure}

\paragraph{Pasadena Multi-View ReID.} Instance labeled trees are ignored in most urban object datasets where they are rather labeled as vegetation. We consider the Pasadena Multi-View ReID \cite{nassar2019simultaneous} dataset that provides labeled instances of different trees acquired in Pasadena, California. The dataset offers approximately 4 Google Street View panoramic views for each tree instance. As shown in Fig.~\ref{fig:pasadena_sample}, a scene of 4 views could contain multiple labeled instances, averaging 2 instances per scene, and other trees that are not instance labeled. There is a total of 6,020 individual tree instances. 

The dataset consists of 6,141 panorama images of size 2048 x 1024 px. In total, there are around 25,061 annotated objects in the dataset. Each annotation includes the object's bounding box values, image geo-coordinates, camera heading, estimate of object distance from camera, ID and geo-coordinate of object. In our experiments, we follow the same data split introduced in \cite{nassar2019simultaneous}, and allocate 4,298 images for training, 921 for validation, and 922 for testing. 

\begin{figure}[!htbp]
\centering
\includegraphics[scale=0.145]{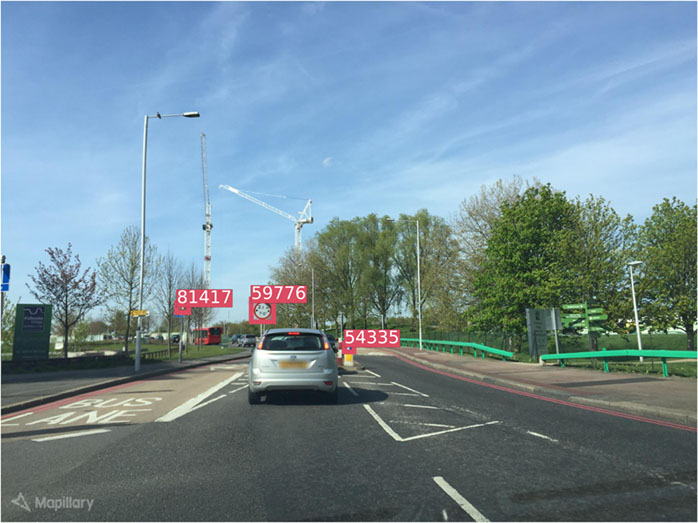}\quad
\includegraphics[scale=0.145]{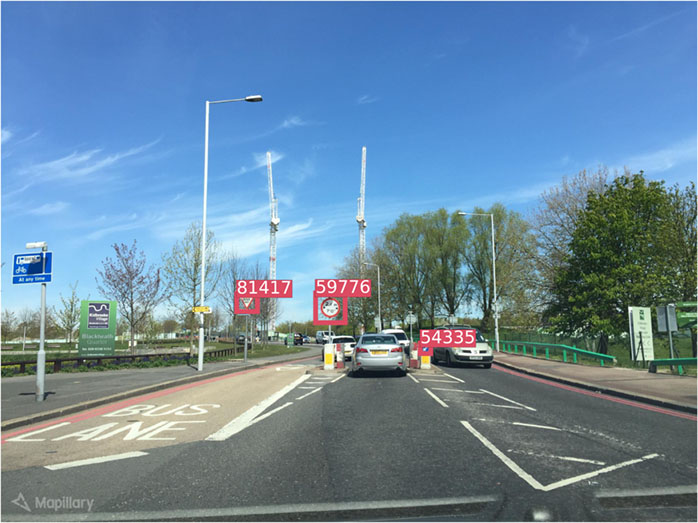}\quad
\medskip
\includegraphics[scale=0.145]{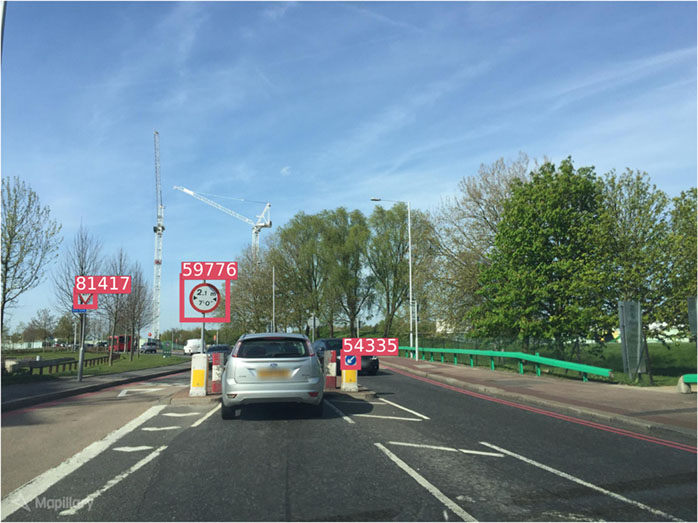}\quad
\includegraphics[scale=0.145]{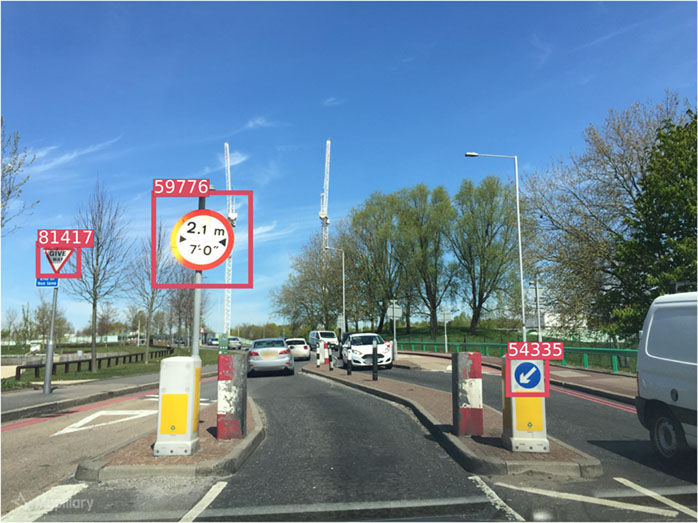}\quad
\caption{Subset of the Mapillary dataset. This scene depicts a dashcam sequence where nearly all signs in the scene are annotated with their instance label (red), and shows strong object scale changes across the different views.}
\label{fig:mapillary_sample}
\end{figure}

\paragraph{Mapillary.} The second dataset we consider is a crowdsourced one provided by Mapillary \footnote{www.mapillary.com}, that is different from  the image segmentation dataset, Mapillary Vistas \cite{neuhold2017mapillary}. Normally, the dataset contains different types of objects, but this subset contains traffic signs only in an area of 2$km^{2}$ London, England. In comparison to the Pasadena Multi-View ReID, the images acquired in this dataset are dominantly captured consecutively by forward-faced cameras mounted on vehicles as shown in \ref{fig:mapillary_sample} with the object mostly facing change in scale with the same viewpoint, with the other instances being from pedestrians' smartphone cameras.

Given its crowdsourced nature, this dataset presents interesting challenges such as images being acquired at different times of the day, various camera sensors as well as image sizes.  There are 31,442 different instances of traffic signs labeled in 74,320 images. An object instance appears on average in 4 images, and there are approximately 2 object instances in each image. Almost entirely, all instances of signs in the image are annotated.
Each object instance in the dataset comes with its bounding box values, object ID, image geo-coordinates, instance geo-coordinates, heading of camera, in which images the object appears, and its height. It is important to note that the object geo-coordinate is attained through 3D Structure From Motion techniques (SFM). In contrast to the Pasadena Multi-View ReID, the objects are much smaller in comparison to trees, and much difficult to capture sideways due to the physical property of signs being thin. On the other hand, the dataset is much larger, thus easing the training process.

\subsection{Implementation Details}
We implemented GeoGraph using PyTorch \cite{paszke2017automatic}. For the object detector, a PyTorch  implementation \cite{EfficientDetSignatrix} of EfficientDet was used. The backbone chosen for the EfficientDet was ``EfficientNet-B5''.
As for the GNN component, we relied on the PyTorch Geometric package \cite{Fey/Lenssen/2019}. The Dropout \cite{srivastava2014dropout} layers are used with a drop probability of $0.2$. The learning rate is set to $0.001$ initially with ADAM \cite{kingma2014adam} as the optimizer. Each epoch takes approximately $45$ minutes during training time on a NVIDIA 1080 Ti GPU.

\subsection{Object Detection \& Re-identification}
\begin{figure}[!htbp]
\centering
\includegraphics[scale=0.255]{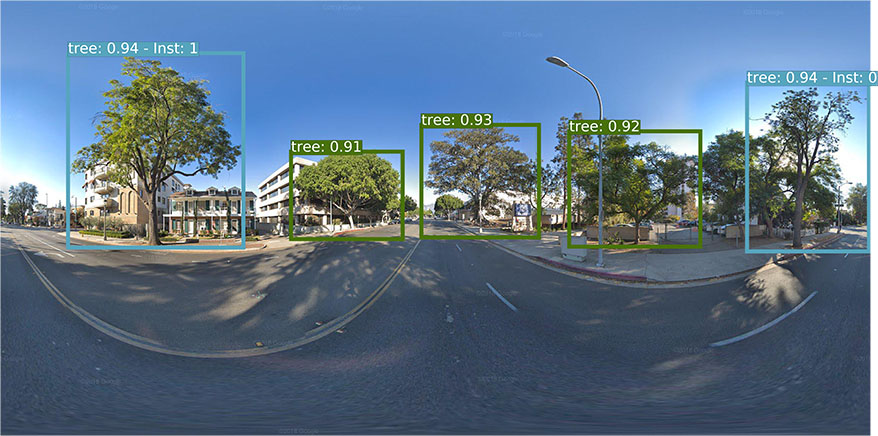}\quad
\includegraphics[scale=0.255]{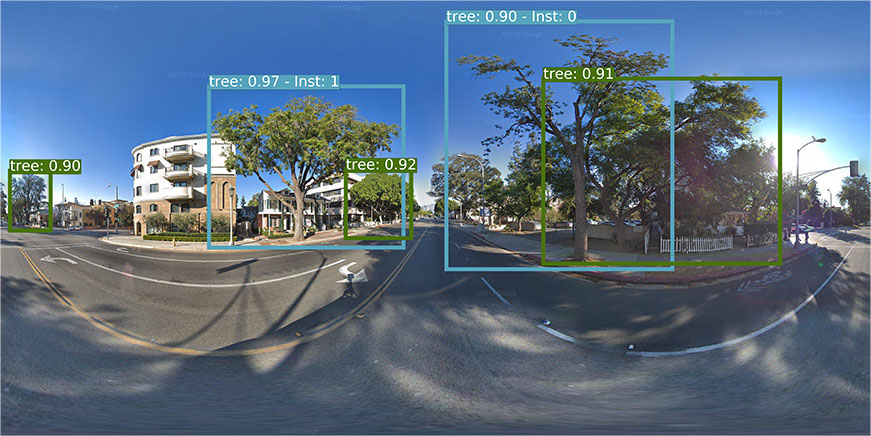}\quad
\includegraphics[scale=0.255]{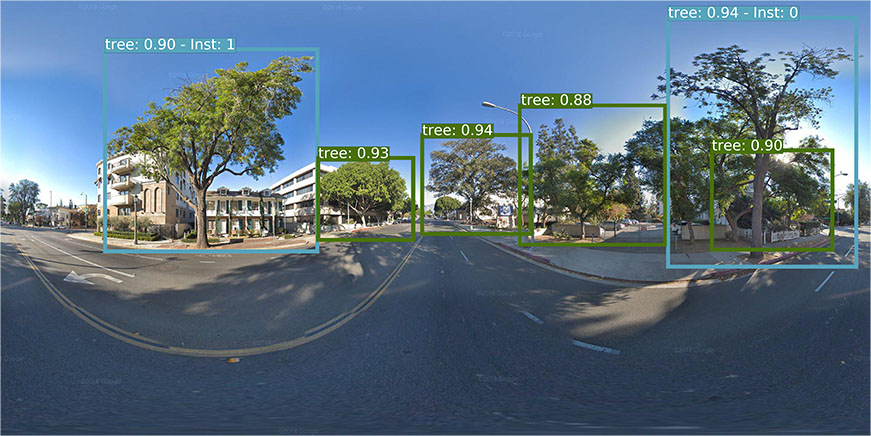}\quad
\includegraphics[scale=0.255]{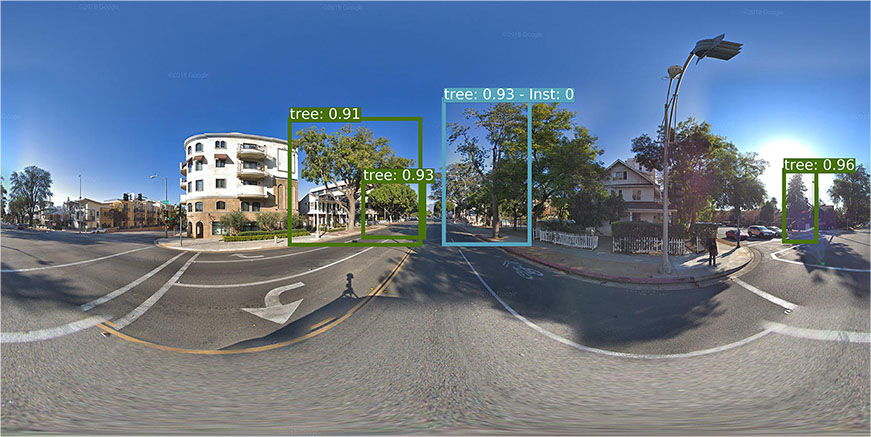}\quad
\caption{Sample results obtained on the Pasadena dataset for multi-view object detection and  re-identification. Trees were correctly detected (green) and further accurately re-identified across different views (cyan) when possible.}
\label{fig:pasadena_sample2}
\end{figure}

\begin{figure}[!htbp]
\centering
\includegraphics[scale=0.26]{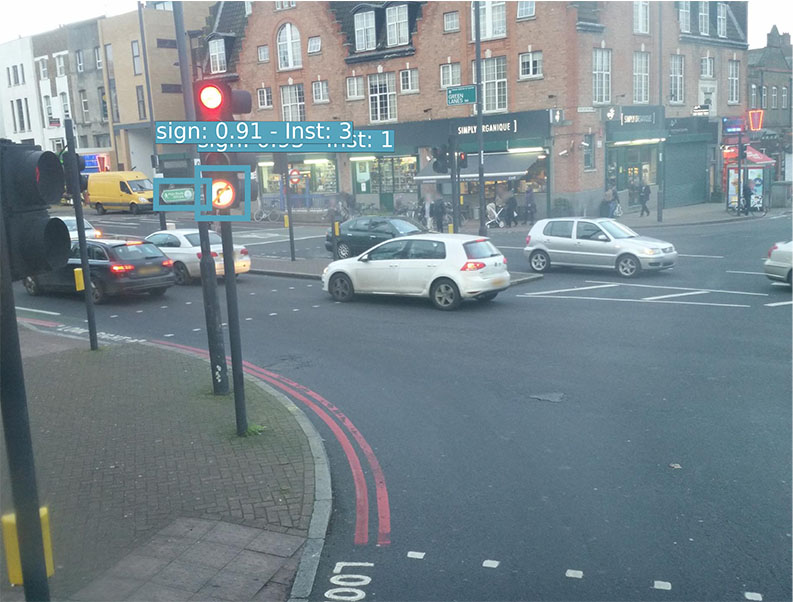}\quad
\includegraphics[scale=0.26]{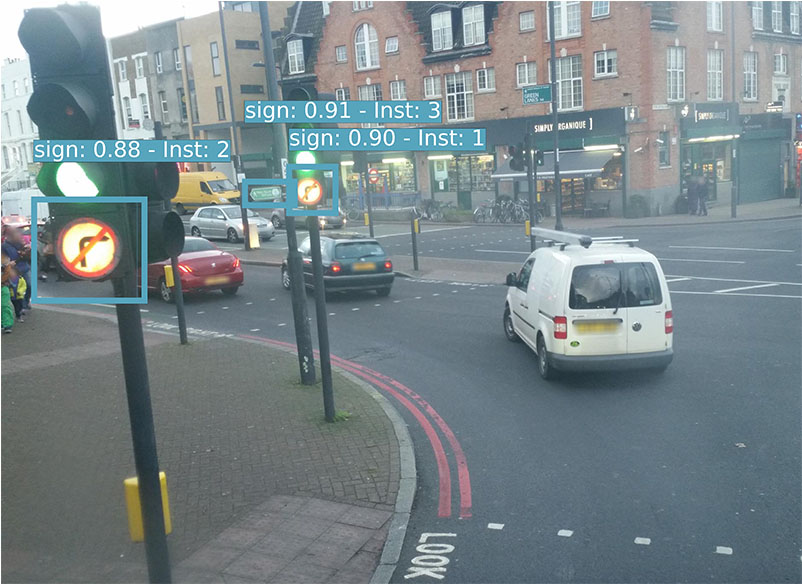}\quad
\caption{Sample results obtained on the Mapillary dataset for multi-view object detection and  re-identification. Here all detected objects (signs) were both detected and further re-identified (cyan) due to the higher similarity between views.}
\label{fig:mapillary_sample2}
\end{figure}

In Tab.~\ref{table:main} we report the results achieved by our method in the different datasets as well as state-of-the-art performance.
The effectiveness of our approach is evaluated by correctly identifying an instance of an object across multiple views, as also visually illustrated in Fig.~\ref{fig:pasadena_sample2} and Fig.~\ref{fig:mapillary_sample2}. Typically, the object detection method influences the re-identification process as a better mean Average Precision (mAP) ensures that the object is fed to the next stage of the pipeline for the re-identification. Moreover, note that the object detection scores for the proposed method does not change between the 2 and 2+ views, this is because the object detection is always carried out on a single image at the time.
Our method outperforms SSD-ReID-Geo \cite{nassar2019simultaneous} for detection mAP by 2.2\% with the Pasadena dataset, and 1.7\% with the Mapillary dataset. GeoGraph improvement can be attributed to the superiority of EfficientDet over SSD. 

Our experiments for re-identification aim at validating whether using graphs with coarse pose information would assist in identifying the same object instance across views. 
The performance of the GNN component is based on whether the detections of the same object across different views have connecting edges. Therefore to calculate our mAP, we consider as true positives correctly predicted edges between a pair of detections with no other edge connections to different objects in the scene. Otherwise we consider them as false positives. In order to ensure a fair comparison with other methods, we perform experiments with a similar number of views. As shown in Tab. \ref{table:main}, increasing the views leads to higher re-identification mAP, with an improvement of 3.2\% for Pasadena and 4.2\% for Mapillary.

\subsection{Geolocalization}
The geo-localization component is evaluated in terms of a Mean Absolute Error (MAE) of the distance, measured in meters, between the predicted geo-coordinate averaged over the different views and the ground truth. The metric chosen to measure the distance is the Haversine distance which is defined as:

\begin{multline}
\label{eq:haversine}
d = 2 R \arcsin\Biggl(\biggl(\sin^2\Bigl(\frac{O_{lat} - G_{lat}}{2}\Bigr) \\
+ \cos(G_{lat}) \cos(O_{lat})
\sin^2\Bigl(\frac{O_{lng} - G_{lng}}{2}\Bigr)\biggr)^{0.5} \Biggr),
\end{multline}

where $O_{lat}, O_{lng}$ represent the detection's predicted geo-coordinates, and $G_{lat}, G_{lng}$ represent the object's ground truth. As reported in Tab.~\ref{table:main}, for the Pasadena dataset, the geo-localization error decreases as the number of views increases (12\% improvement). As for Mapillary, we report again lower error with our GeoGraph and 6 views w.r.t. SSD-ReID-Geo \cite{nassar2019simultaneous} (3.4\% improvement). However, we surprisingly did not outperform GeoGraph results achieved with only 2 views. This  is probably be due to the way the ground truth of the Mapillary dataset is acquired (see Sec. \ref{sec:datasets}), and adding more views brings a lot of noise in the representation of a scene and of the objects it contains.

\begin{table*}
\begin{center}
\caption{Quantitative assessment of our GeoGraph framework and related work on object detection, re-identification, and geo-localization tasks.}
\label{table:main}
\setlength{\tabcolsep}{4pt}
\begin{tabular}{cccrcc}
\hline\noalign{\smallskip}
Method & \# Views & Dataset & \multicolumn{1}{c}{Detection} & Re-ID  & Geo-localization\\
& & & \multicolumn{1}{c}{mAP} & mAP & error (m)\\
\noalign{\smallskip}
\hline
\noalign{\smallskip}
SSD-ReID-Geo \cite{nassar2019simultaneous} & 2 & Pasadena & 0.682 & 0.731 & 3.13\\
Our GeoGraph & 2 & Pasadena & \multirow{2}{*}{\Big\} 0.742} & \textbf{0.754} & \textbf{2.94}\\ 
Our GeoGraph & 4 & Pasadena &  & \textbf{0.763} & \textbf{2.75}\\
\hline%
SSD-ReID-Geo \cite{nassar2019simultaneous} & 2 & Mapillary & 0.902 & 0.882 & 4.36\\
Our GeoGraph & 2  & Mapillary & \multirow{2}{*}{\Big\} 0.919} & \textbf{0.902} & \textbf{3.88}\\
Our GeoGraph & 6 & Mapillary &  & \textbf{0.924} & 4.21\\ 
\hline
\end{tabular}
\end{center}
\end{table*}
\setlength{\tabcolsep}{1.4pt}


\subsection{Ablation Studies}
Since the graphs are generated during training are created online, we assess this component separately to be able to evaluate its effect. In these experiments, we build the graph from our ground-truth by generating CNN features from the image crops of multi-view images, which served as our node features. Using the labeled instances, edges were associated between the different instances of image crops across the different views. Throughout this experiment, the same settings of the GNN were used as mentioned in Sec. \ref{method:gnn}. We compare our method to a siamese CNN with a ResNet50 backbone trained with a contrastive loss that classified whether the two crops of the object are similar or not.
\begin{table}
\begin{center}
\caption{Results for our graph matching method component evaluated with bypassing the object detector, and using image crops. We show a comparison between a Siamese CNN, a GNN based on CNN features and a GNN based on CNN features and camera metadata information to classify if pairs of objects are the same or not.}
\label{table:ablation_graph}
\setlength{\tabcolsep}{4pt}
\begin{tabular}{llc}
\hline\noalign{\smallskip}
Method & Dataset & F1-Score\\
\noalign{\smallskip}
\hline
\noalign{\smallskip}
Siamese CNN & Pasadena & 0.509\\
GNN  & Pasadena & 0.601 \\
GNN-Geo & Pasadena & \textbf{0.640} \\ 
\hline%
Siamese CNN & Mapillary & 0.721\\
GNN & Mapillary & 0.823\\
GNN-Geo & Mapillary & \textbf{0.873}\\
\hline
\end{tabular}
\end{center}
\end{table}

We can observe from results reported in Tab. \ref{table:ablation_graph} that adding geometric cues consistently helped to improve re-identification. Through different examination with different forms of the graph construction, we have found out that creating edges between nodes across multiple views but also within the same image in a fully connected dense manner led to better results than in the case where edges between nodes of the same image (i.e. detections within same image) are ignored. This can be explained by the fact that node aggregation performs better with fully connected graphs.

\section{Conclusion}

In this paper we have tackled the problems of object detection, re-identification across multiple views and geo-localization with a unified, end-to-end learnable framework. We propose a method that integrates both image and pose information and use a GNN to perform object re-identification. The advantage of the our GNN over standard Siamese CNN is the ability to deal with any number of views and be the computational efficiency. Experiments conducted on two public datasets have shown the relevance of our GeoGraph framework, which achieves high detection accuracy together with low geo-localization error. Furthermore, our approach is robust to occlusion, neighboring objects of similar appearance, and severe changes in viewpoints.

The proposed framework could be improved if based on a better geo-localization component. Furthermore, using the proposals generated from each view could improve the quality of the object detection step. Finally, while only street view images were used in our paper, our GeoGraph framework is compatible with other types of images. We thus would like to combine these ground-level views with aerial views in order to improve the overall performance and to make the best of multiple viewpoints following \cite{lefevre2017toward}.

{\small
\bibliographystyle{ieee_fullname}
\bibliography{egbib}
}

\end{document}